\pdfoutput=1

\documentclass[11pt]{article}
\usepackage{emnlp2021}
\usepackage{times}
\usepackage{latexsym}
\usepackage{cleveref}

\usepackage{xcolor}

\usepackage{graphicx}
\usepackage{lscape}
\usepackage{subfig}

\graphicspath{ {img/} }
\usepackage{latexsym}

\usepackage{tabularx}
\usepackage{booktabs}
\usepackage{multirow}
\usepackage{caption}
\usepackage{siunitx}
\usepackage{amssymb}
\usepackage[flushleft]{threeparttable}
\usepackage{url}
\usepackage{floatrow}
\usepackage{wrapfig}
\usepackage{microtype}

\title{Detecting Action Reason\oana{intent?} in Lifestyle Vlogs}
\title{{\sc WhyAct:} Identifying Action Reasons in Lifestyle Vlogs}

\author{Oana Ignat \and Santiago Castro \and  Hanwen Miao \and Weiji Li \and Rada Mihalcea \\
Computer Science and Engineering \\
University of Michigan\\
{\tt \{oignat,sacastro,hwmiao,weijili,mihalcea\}@umich.edu}
}

\begin{document}

\maketitle
\begin{figure*}
    \centering
    \includegraphics[width=0.97\textwidth]{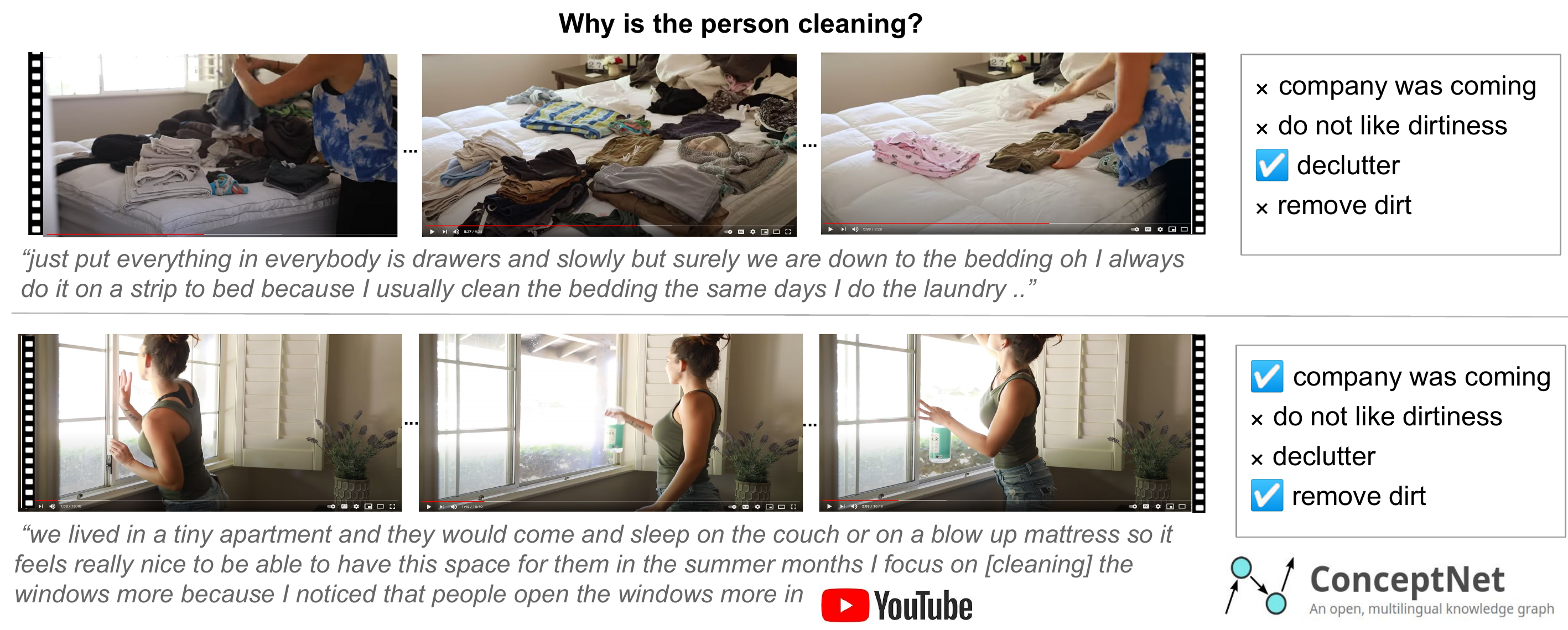}
    \caption{Overview of our task: automatic identification of action reasons in online videos. The reasons for {\it cleaning} change based on the visual and textual (video transcript) context. The videos are selected from YouTube, and the actions together with their reasons are obtained from the ConceptNet \cite{Speer2017ConceptNet5A} knowledge graph which we supplement with crowdsourced reasons. The figure shows two examples from our {\sc WhyAct} dataset.}
    \label{fig:example_clean_fig1}
\end{figure*}

\begin{abstract}
We aim to automatically identify human action reasons in online videos. We focus on the widespread genre of lifestyle vlogs, in which people perform actions while verbally describing them. We introduce and make publicly available the {\sc WhyAct} dataset, consisting of 1,077 visual actions manually annotated with their reasons. We describe a multimodal model that leverages visual and textual information to automatically infer the reasons corresponding to an action presented in the video. 
\end{abstract}


\section{Introduction}
Significant research effort has been recently devoted to the task of action recognition 
\cite{ Carreira2017QuoVA, Shou2017CDCCN, Tran2018ACL, Chao2018RethinkingTF, Girdhar2019VideoAT, Feichtenhofer2019SlowFastNF}.
Action recognition works well when applied to well defined/constrained scenarios, such as people following scripts and instructions \cite{Sigurdsson2016HollywoodIH, Miech2019HowTo100MLA, Tang2019COINAL}, performing sports \cite{Soomro2012UCF101AD, Karpathy2014LargeScaleVC} or cooking \cite{Rohrbach2012ADF, Damen2018EPICKITCHENS, Damen2020RESCALING, Zhou2018TowardsAL}. At the same time however, action recognition is limited and error-prone  once the application space is opened to everyday life. This indicates that current action recognition systems rely mostly on pattern memorization and do not effectively understand the action, which makes them fragile and unable to adapt to new settings \cite{Sigurdsson2017WhatAA, Kong2018HumanAR}. Research on how to improve action recognition in videos \cite{Sigurdsson2017WhatAA} shows that recognition systems for actions with known intent have a significant increase in performance, as 
knowing the reason for performing an action is an important step for understanding that action \cite{Tosi1991ATO, Gilovich2002HeuristicsAB}.

In contrast to action recognition, action causal reasoning research is just emerging in computational applications \cite{Vondrick2016PredictingMO, Yeo2018VisualCO, Zhang2020LearningCC,Fang2020Video2CommonsenseGC}. Causal reasoning has direct applications on many real-life settings, for instance to understand the consequences of events (e.g., if ``there is clutter,'' ``cleaning'' is required), or to enable social reasoning (e.g., when ``guests are expected,'' ``cleaning'' may be needed -- see \Cref{fig:example_clean_fig1}). 
Most of the work to date on causal systems has relied on the use of semantic parsers to identify reasons \cite{He2017DeepSR}, however this approach does not work well on more realistic every-day settings. 
As an example, consider the statement ``This is a mess and my friends are coming over. I need to start cleaning.'' Current causal systems are unable to identify ``this is a mess'' and ``friends are coming over'' as reasons, and are thus failing to use them as context for understanding the action of ``cleaning.''


In this paper,  we propose the task of multimodal action reason identification in  everyday life scenarios. We collect a dataset of lifestyle vlogs from YouTube that reflect daily scenarios and are currently very challenging for systems to solve. Vloggers freely express themselves while performing most common everyday activities such as cleaning, eating, cooking, writing and others. 
Lifestyle vlogs present a person's everyday routine: 
the vlogger visually records the activities they perform during a normal day and verbally express their intentions and feelings about those activities. 
Because of these characteristics, lifestyle vlogs are a rich data source for an in depth study of human actions and the reasons behind them. 


The paper makes four main contributions. First, we formalize the new task of multimodal action reason identification in online vlogs. Second, we introduce a new dataset, {\sc WhyAct}, consisting of 1,077 (action, context, reasons) tuples manually labeled in online vlogs, covering 24 actions and their reasons drawn from ConceptNet as well as crowdsourcing contributions. Third, we propose several models to solve the task of human action reason identification, consisting of  single-modalities models based on the visual content and vlog transcripts, as well as a multimodal model using a fill-in-the-blanks strategy. Finally, we also present an analysis of our new dataset, which leads to rich avenues for future work for improving the tasks of reason identification and ultimately action recognition in online videos. 

\section{Related Work}
There are three areas of research related to our work: identifying action motivation, commonsense knowledge acquisition, and web supervision. 
\paragraph{Identifying Action Motivation.}
The research most closely related to our paper is the work that introduced the task of predicting motivations of actions by leveraging text \cite{Vondrick2016PredictingMO}. Their method was applied to  images from the COCO dataset \cite{Lin2014MicrosoftCC}, while ours is focused on videos from YouTube.  
Other work on human action causality in the visual domain \cite{Yeo2018VisualCO, Zhang2020LearningCC} relies on object detection and automatic image captioning as a way to represent videos and analyze visual causal relations.
Research has also been carried out on detecting the intentions of human actions  \cite{Pezzelle2020BeDT}; the task definition differs from ours, however, as their goal is to automatically choose the correct action for a given image and intention. 
Other related work includes
\cite{Synakowski2020AddingKT}, a vision-based classification model between intentional and non-intentional actions and Intentonomy \cite{Jia2020IntentonomyAD}, a dataset on human intent behind images on Instagram.

\paragraph{Commonsense Knowledge Acquisition.}
Research on commonsense knowledge often relies on  textual knowledge bases such as
ConceptNet \cite{Speer2017ConceptNet5A}, ATOMIC \cite{Sap2019ATOMICAA}, COMET-ATOMIC 2020 \cite{Hwang2020COMETATOMIC2O}, and more recently GLUCOSE \cite{Mostafazadeh2020GLUCOSEGA}.


Recently, several of these textual knowledge bases have also been used for visual applications, to create more complex multimodal datasets and models \cite{Park2020VisualCOMETRA, Fang2020Video2CommonsenseGC, Song2020KVLBERTKE}.
VisualCOMET \cite{Park2020VisualCOMETRA} is a dataset for visual commonsense reasoning tasks to predict events that might have happened before a given event, events that might happen next, as well as people intents at a given point in time. Their dataset is built on top of VCR \cite{zellers2019vcr}, which consists of images of multiple people and activities.
Video2Commonsense \cite{Fang2020Video2CommonsenseGC} uses ATOMIC to extract from an input video a list of intentions that are provided as input to a system that generates video captions, as well as three types of commonsense descriptions (intention, effect, attribute).
KVL-BERT \cite{Song2020KVLBERTKE} proposes a knowledge enhanced cross-modal BERT model by introducing entities extracted from ConceptNet \cite{Speer2017ConceptNet5A} into the input sentences, followed by testing their visual question answering model on the VCR benchmark \cite{zellers2019vcr}.
Unlike previous work that broadly addresses commonsense relations, we focus on the extraction and analysis of action reasons, which allows us to gain deeper insights for this relation type.



\paragraph{Webly-Supervised Learning.} 
The space of current commonsense inference systems is often limited to one dataset at a time, e.g., COCO \cite{Lin2014MicrosoftCC}, VCR \cite{zellers2019vcr}, MSR-VTT \cite{Xu2016MSRVTTAL}. In our work, we ask commonsense questions in the context of rich, unlimited, constantly evolving online videos from YouTube.

Previous work  has leveraged webly-labeled data for the purpose of identifying commonsense knowledge. One of the most extensive efforts is NELL (Never Ending Language Learner) \cite{Mitchell2015NeverEndingL}, a system that learns everyday knowledge by crawling the web, reading documents and analysing their linguistic patterns. A closely related effort is  NEIL (Never Ending Image Learner), which learns commonsense knowledge from images on the web \cite{Chen2013Neil}. Large scale video datasets \cite{Miech2019HowTo100MLA} on instructional videos and lifestyle vlogs \cite{Fouhey2018FromLV, Ignat2019IdentifyingVA} are other examples of web supervision. The latter are similar to our work as they analyse online vlogs, but unlike our work, their focus is on action detection and not on the reasons behind  actions.

\section{Data Collection and Annotation}
In order to develop and test models for recognizing reasons for human actions in videos, we need a manually annotated dataset. This section describes the {\sc WhyAct} dataset of action reasons.

\subsection{Data Collection}
We start by compiling a set of lifestyle videos
from YouTube, consisting of people performing their daily routine activities, such as cleaning, cooking, studying, relaxing, and others.
We build a data gathering pipeline to automatically extract and filter videos and their transcripts. 

We select five YouTube channels and download all the videos and their transcripts. The channels are selected to have good quality videos with automatically generated transcripts containing detailed verbal descriptions of the actions depicted. 
An analysis of the videos indicates that both the textual and visual information are rich sources for describing not only the actions, but why the actions in the videos are undertaken (action reasons). We present qualitative and quantitative analyses of our data in section \ref{data_analysis}.

We also collect a set of human actions and their reasons from ConceptNet \cite{Speer2017ConceptNet5A}. Actions include verbs such as: \textit{clean}, \textit{write}, \textit{eat}, and other verbs describing everyday activities. The actions are selected based on how many reasons are provided in ConceptNet and how likely they are to appear in our collected videos. For example, the action of \textit{cleaning} is likely to appear in the vlog data, while the action of \textit{yawning} is not.

\subsection{Data Pre-processing}
After collecting the videos, actions and their corresponding reasons, the following data pre-processing steps are applied.

\begin{table}
    \centering
    \begin{tabular}{l c }
        \toprule
        Initial  &  9,759 \\
        Actions with reasons in ConceptNet & 139 \\
        Actions with at least 3 reasons in CN & 102 \\
        Actions with at least 25 video-clips & 25 \\
    \bottomrule 
    \end{tabular}
    \caption{Statistics for number of collected actions at each stage of data filtering.}
    \label{tab:reasons_stats}
\end{table}

\paragraph{Action and Reason Filtering.}
From ConceptNet, we select actions that contain at least three reasons. The reasons in ConceptNet are marked by the ``motivated by`` relation. We further filter out those actions that appear less than 25 times in our video dataset, in order to assure that each action has a significant number of instances. 

We find that the reasons from ConceptNet are often very similar to each other, and thus easy to confound. For example, the reasons for the action \textit{clean} are:
``dirty", ``remove dirt", ``don’t like dirtiness", ``there dust", ``dirtiness unpleasant", ``dirt can make ill", ``things cleaner", ``messy", ``company was coming".
To address this issue, we apply  agglomerative clustering \cite{Murtagh2014WardsHA} to group similar actions together. For instance, for the action \textit{clean}, the following clusters are produced: [``dirty", ``remove dirt", ``there dust", ``things cleaner"], [``don't like dirtiness", ``dirtiness unpleasant", ``dirt can make ill"], [``messy"], [``company was coming"]. Next, we manually select the most representative and clear reason from each cluster. We also correct any spelling mistakes and rename the reasons that are either too general or unclear (e.g., we rename ``messy" to ``declutter").
Finally, after the clustering and processing steps, we filter out all the actions that contain less than three reasons.

We show the statistics before and after the additive filtering steps in \Cref{tab:reasons_stats}.

\paragraph{Transcript Filtering.}
We want transcripts that reflect the reasons for performing one or more actions shown in the video. However,
the majority of the transcripts contain mainly verbal descriptions of the action, which are not always helpful in determining their reason. We therefore implement a method to select candidate transcript sequences that contain at least one causal relation related to the actions shown in the video.

We start by automatically splitting the transcripts into sentences using spaCy \cite{spacy}. Next, we select the sentences with at least one action from the final list of actions we collected from ConceptNet (see the previous section). For each selected sentence, we also collect its context consisting of the sentences before and after. We do this in order to increase the search space for the reasons for the actions mentioned in the selected sentences. 

We want to keep the sentences that contain action reasons. We tried multiple methods to automatically determine the sentences more likely to include  causal relations using Semantic Role Labeling (SRL) \cite{Ouchi2018ASS}, Open Information Extraction (OpenIE) \cite{Angeli2015LeveragingLS} and searching for causal markers. We found that SRL and OpenIE do not work well on our data, likely due to the fact that the transcripts are more noisy than the datasets these models were trained on. Most of the language in the transcripts does not follow simple patterns such as  ``I clean because it is dirty." Instead, the language consists of natural everyday speech such as ``Look at how dirty this is, I think I should clean it." 

We find that a strategy sufficient for our purposes is to search for causal markers such as ``because", ``since", ``so that is why", ``thus",  ``therefore" in the sentence and the context, and constrain the distance between the actions and the markers to be less than 15 words -- a threshold identified on development data.
We thus keep all the transcript sentences and their context that contain at least one action and a causal marker within a distance of less than the threshold of 15 words.

\paragraph{Video Filtering.}
As transcripts are temporally aligned with videos, we can obtain meaningful video clips related to the narration. We extract video clips corresponding to the sentences selected from transcripts (described in the section above).

We want video clips that show why the actions are being performed. Although there can be many actions along with reasons in the transcript, if they are not depicted in the video, we cannot leverage the video information in our task.  Videos with low movement tend to show people sitting in front of the camera, describing their routine, but not performing the action they are talking about. 
We therefore remove clips that do not contain enough movement. We sample one out of every one hundred frames of the clip, and compute the 2D correlation coefficient between these sampled frames. If the median of the obtained values is greater than a certain threshold (0.8, selected on the development data), we filter out the clip. 
We also remove video-clips that are shorter than 10 seconds and longer than 3 minutes.

\begin{figure}
\scalebox{0.9}{
    \centering
    \includegraphics[width=\textwidth]{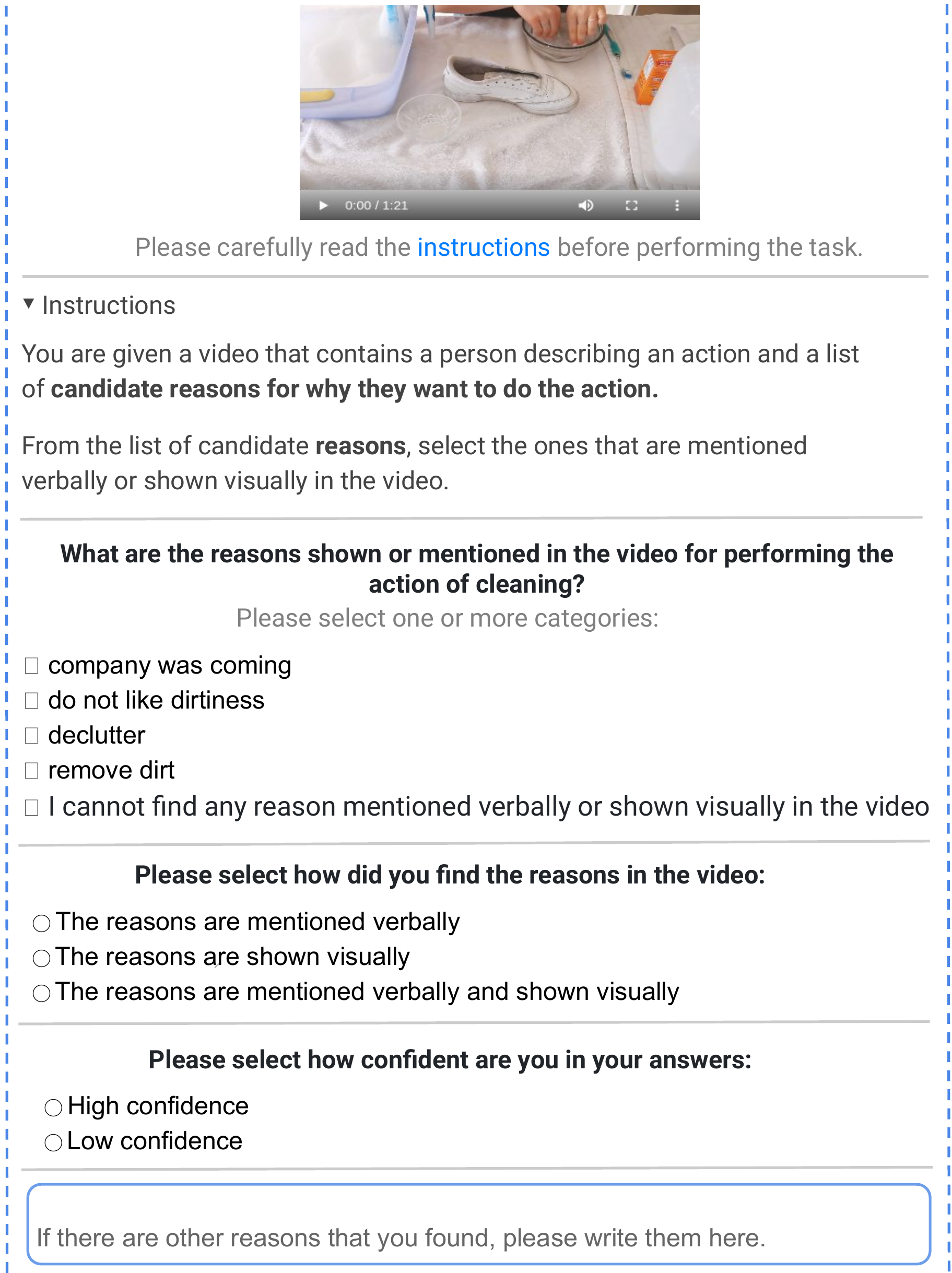}
    }
    \caption{Instructions for the annotators.}
    \label{fig:amt_gui}
\end{figure}

\subsection{Data Annotation}

The resulting (video clip, action, reasons) tuples are annotated with the help of Amazon Mechanical Turk (AMT) workers. They are asked to identify: (1) what are the reasons shown or mentioned in the video clip for performing a given action; (2) how are the reasons  identified in the video: are they mentioned verbally, shown visually, or both; (3) whether there are other reasons other than the ones provided; (4) how confident the annotator is in their response. The guidelines and interface for annotations are shown in \Cref{fig:amt_gui}. In addition to the guidelines, we also provide the annotators with a series of examples of completed assignments with explanations for why the answers were selected. We present them in the supplemental material in \Cref{fig:supp_amt}.

We add new action reasons from the ones added by the annotators if they repeat at least three times in the collected answers and are not similar to the ones already existing.

Each assignment is completed by three different master annotators. We compute the agreement between the annotators using Fleiss Kappa \cite{Fleiss1971MeasuringNS}  and we obtain 0.6, which indicates a moderate agreement.
Because the annotators can select multiple reasons, the agreement is computed per reason and then averaged. 

We also analyse how confident the workers are in their answers: for each video, we take the confidence selected by the majority of workers: out of 1,077 videos, in 890 videos the majority of workers are highly confident.

\Cref{tab:statistics} shows statistics for our final dataset of video-clips and actions annotated with their reasons. \Cref{fig:example_clean_fig1} shows a sample video and transcript, with annotations. Additional examples of annotated actions and their reasons can be seen in the supplemental material in \Cref{fig:other_examples}.

\begin{table}
    \centering
    \scalebox{0.9}{
    \begin{tabular}{l r}
    \toprule
    Video-clips & 1,077  \\
    Video hours & 107.3 \\
    Transcript words & 109,711 \\
    Actions & 24 \\
    Reasons & 166 \\
    \bottomrule
    \end{tabular}
    }
    \caption{Data statistics.}
    \label{tab:statistics}
\end{table}

\begin{table}
    \centering
    \scalebox{0.9}{
    \begin{tabular}{l c c c}
    \hline
     & Test & Development\\
        \midrule
        Actions & 24 & 24 \\
        Reasons & 166 & 166 \\
        Video-clips & 853 & 224\\
    \hline 
    \end{tabular}
    }
    \caption{Statistics for the experimental data split. The methods we run are unsupervised with fine-tuning on development set.}
    \label{tab:train-test-eval-split}
\end{table}

\section{Identifying Causal Relations in Vlogs}

Given a video, an action, and a list of candidate action reasons, our goal is to determine the reasons mentioned or shown in the video. We develop a multimodal model that leverages both visual and textual information, and we compare its performance with several single-modality baselines.

The models we develop are unsupervised in that we are not learning any task-specific information from a training dataset. We use a validation set only to tune the hyper-parameters of the models.


\subsection{Data Processing and Representation}\label{data_proc}

\paragraph{Textual Representations.}
To represent the textual data -- transcripts and candidate reasons -- we use sentence embeddings computed using the pre-trained model Sentence-BERT \cite{reimers-2019-sentence-bert}.

\paragraph{Video Representations.}
In order to tie together the causal relations, both the textual, and the visual information, we represent the video as a bag of object labels and a collection of video captions.
For object detection we use Detectron2 \cite{wu2019detectron2}, a state-of-the-art object detection algorithm. 

We generate automatic captions for the videos using a state-of-the-art dense captioning model \cite{BMT_Iashin_2020}. The input to the model are visual features extracted from I3D model pre-trained on Kinetics \cite{Carreira2017QuoVA}, audio features extracted with VGGish model \cite{45611} pre-trained on YouTube-8M \cite{AbuElHaija2016YouTube8MAL} and caption tokens using GloVe \cite{Pennington2014GloveGV}.

\subsection{Baselines}
Using the representations described in \Cref{data_proc}, we implement several textual and visual models. 


\begin{figure*}[h]
\scalebox{1}{
    \centering
    \includegraphics[width=0.95\textwidth]{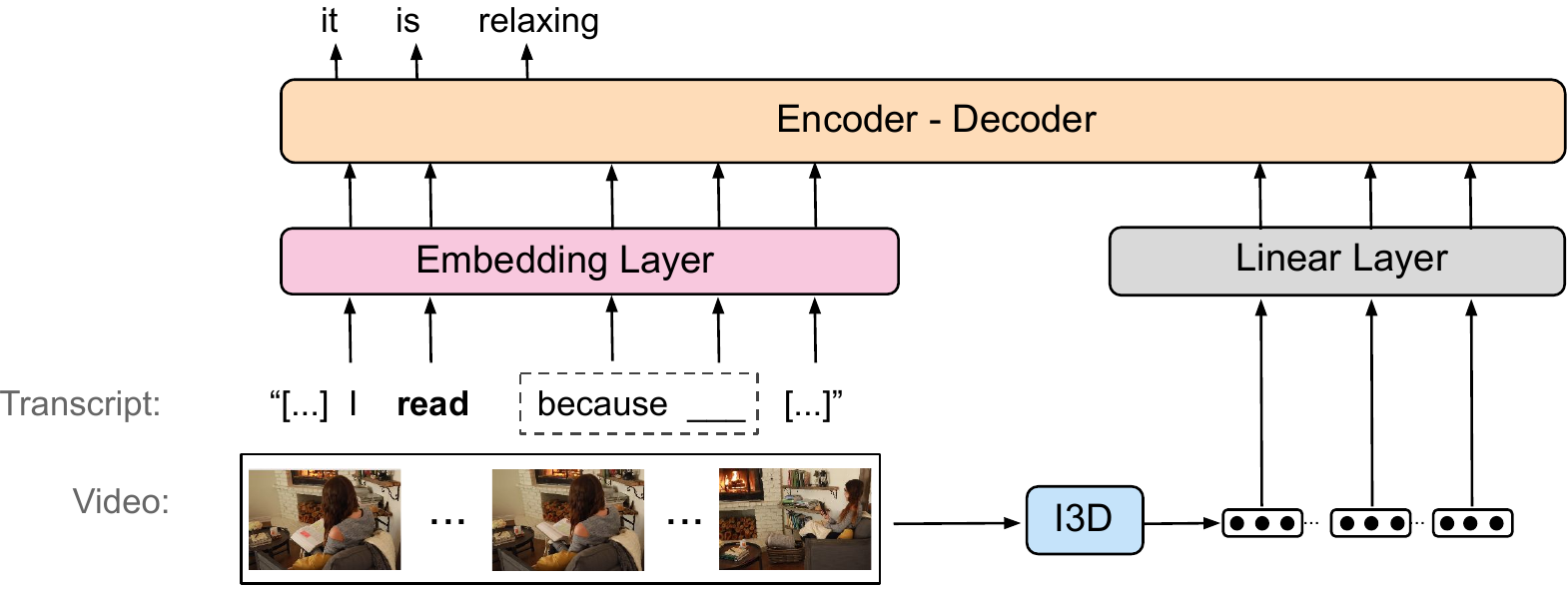}
    }
    \caption{Overview architecture of our Multimodal Fill-in-the-blanks model. The span of text ``because \_\_\_\_\_" is introduced in the video transcript, after the appearance of the action. This forces the T5 model to generate the words missing in the blanks. We then compute the probability of each potential reason and take as positive those that pass a threshold.}%
    \label{fig:main_model}
\end{figure*}

\subsubsection{Textual Similarity}
Given an action, a video transcript associated with the action, and a list of the candidate action reasons, we compute the cosine similarity between the textual representations of the transcript and all the candidate reasons. We predict as correct those reasons that have a cosine similarity with the transcript greater than a threshold of 0.1. The threshold is fine-tuned on development data. 

Because the transcript might contain information that may be unrelated to the action described or its reasons, we also develop a second version of this baseline. When computing the similarity, instead of using the whole transcript, we select only the part of the transcript that is in the vicinity of the causal markers (before and after a fixed number words, fine-tuned on development data). 

\subsubsection{Natural Language Inference (NLI)}
We use a pre-trained NLI model \cite{Yin2019BenchmarkingZT} as a zero-shot sequence classifier. The NLI model is pre-trained on the Multi-Genre Natural Language Inference (MultiNLI) corpus \cite{N18-1101}, a collection of sentence pairs annotated with textual entailment information.

The method works by posing the sequence to be classified as the NLI premise and constructing a hypothesis from each candidate label: given the transcript as a premise and the list of reasons as the hypotheses, each reason will receive a score that reflects the probability of entailment.
For example, if we want to evaluate whether the label ``declutter" is a reason for the action ``cleaning", we construct the hypothesis ``The reason for cleaning is declutter."

We use a threshold of 0.8 fine-tuned on the development data to filter the reasons that have a high entailment score with the transcript. 

\paragraph{Bag of Objects.}
We replace the transcript in the premise with a list of object labels  detected from the video. The objects are detected using the Detectron2 model \cite{wu2019detectron2} on each video frame, at 1fps. We select only the objects that pass a confidence score of 0.7.

\paragraph{Automatic Video Captioning.}
We replace the transcript in the premise with a list of video captions  detected using the Bi-modal Transformer for Dense Video Captioning model \cite{BMT_Iashin_2020}. The video captioning  model generates captions for several time slots. We further filter the generated captions to remove redundant captions: if a time slot is heavily overlapped or even covered by another time slot, we only keep the caption of the longer time slot. We find that captions of longer time slots are also more informative and accurate compared to captions of shorter time slots.

\subsection{Multimodal Model}

To leverage information from both the visual and linguistic modalities, we propose a new model that recasts our task as a Cloze task, and attempts to identify the action reasons by performing a fill-in-the-blanks prediction, similarly to~\citet{fitb} that proposes to fill blanks corresponding to noun phrases in descriptions based on video clips content. Specifically, after each action mention for which we want to identify the reason, we add the text ``because \_\_\_\_\_.'' For instance, ``I clean the windows'' is replaced by ``I clean the windows because \_\_\_\_\_''. We train a language model to compute the likelihood of filling in the blank with each of the candidate reasons. For this purpose, we use T5~\cite{t5}, an encoder-decoder transformer~\cite{transformer} pre-trained model, to fill in blanks with  text.

To incorporate the visual data, we first obtain Kinetics-pre-trained I3D~\cite{Carreira2017QuoVA} RGB features at 25fps (the average pooling layer). We input the features to the T5 encoder after the transcript text tokens. The text input is passed through an embedding layer (as in T5), while the video features are passed through a linear layer. Since T5 was not trained with this kind of input, we fine-tune it on unlabeled data from the same source, without including data that contains the causal marker ``because''. Note this also helps the model specialize on filling-in-the-blank with reasons. Finally, we fine-tune the model on the development data. We obtain the reasons for an action by computing the likelihood of the potential ones and taking the ones that pass a threshold selected based on the development data.
The model architecture is shown in \Cref{fig:main_model}.

We also use our fill-in-the-blanks model in a single modality mode, where we apply it only on the transcript. 

\section{Evaluation}

We consider as gold standard the labels selected by the majority of workers (at least two out of three workers). 

For our experiments, we split the data across video-clips: 20\% development and 80\% test (see \Cref{tab:train-test-eval-split} for a breakdown of actions, reasons and video-clips in each set).
We evaluate our systems as follows. For each action and corresponding video-clip, we compute the Accuracy, Precision, Recall and F1 scores between the gold standard and predicted labels. We then compute the average of the scores across actions. Because the annotated data is unbalanced (in average, 2 out of 6 candidate reasons per instance are selected as gold standard), the most representative metric is F1 score. The average results are shown in \Cref{tab:eval}. The results also vary by action: the F1 scores for each action, of the best performing method, are shown in the supplemental material in \Cref{fig:distrib_f1}.

Experiments on {\sc WhyAct} reveal that both textual and visual modalities contribute to solving the task. The results demonstrate that the task is challenging and there is room for improvement for future work models. 

Selecting the most frequent reason for each action on test data achieves on average an F1 of 40.64, with a wide variation ranging from a very low F1 for the action ``writing'' (7.66 F1) to a high F1 for the action ``cleaning" (55.42 F1).
Note however that the ``most frequent reason'' model makes use of data distributions that our models do not use (because our models are not trained). Furthermore, we believe that it is expected that for certain actions the distribution of reasons is unbalanced, as in everyday life there are action reasons much more common than others (e.g. for ``cleaning", ``remove dirt" is a more common/frequent reason than ``company was coming").

\begin{table*}[!ht]
    \begin{tabular}{l|l|c c c c c c}
    \toprule
    Method & Input & Accuracy & Precision &  Recall &  F1\\
        \midrule
        \midrule
        \multicolumn{6}{c}{\sc Baselines} \\
        \midrule
        \multirow{2}{5pt}{Cosine similarity} & Transcript & 57.70 & 31.39 & 55.94 & 37.64 \\
        & Causal relations from transcript & 50.85 & 30.40 & 68.91 & 39.73\\
        \midrule
        \midrule
        \multicolumn{6}{c}{\sc Single Modality Models}\\ \midrule 
\multirow{2}{5pt}{Natural Language Inference} & Transcript &  {\bf 68.41} & {\bf 41.90} & 48.01 & 40.78 \\
         & Video object labels & 54.49 & 31.70 & 59.93 & 36.79 \\
         & Video dense captions & 49.18 & 29.54 & 68.47 & 37.40 \\
         & Video object labels \& dense captions & 36.93 & 27.34 & 87.97 & 39.11 \\
         \midrule
         Fill-in-the-blanks  & Transcript & 44.04 & 30.70 & 87.10 & {\bf 43.59} \\
        \midrule
        \midrule
        \multicolumn{6}{c}{\sc Multimodal Neural Models} \\
        \midrule
        \midrule
        Fill-in-the-blanks  & Video \& Transcript & 32.6 & 27.56 & {\bf 94.76} & 41.11 \\
          \bottomrule
    \end{tabular}
    \caption{Results from our models on test data.}
    \label{tab:eval}
    
\end{table*}

\begin{figure}[h]
\scalebox{1.1}{
    \centering
    \includegraphics[width=\textwidth]{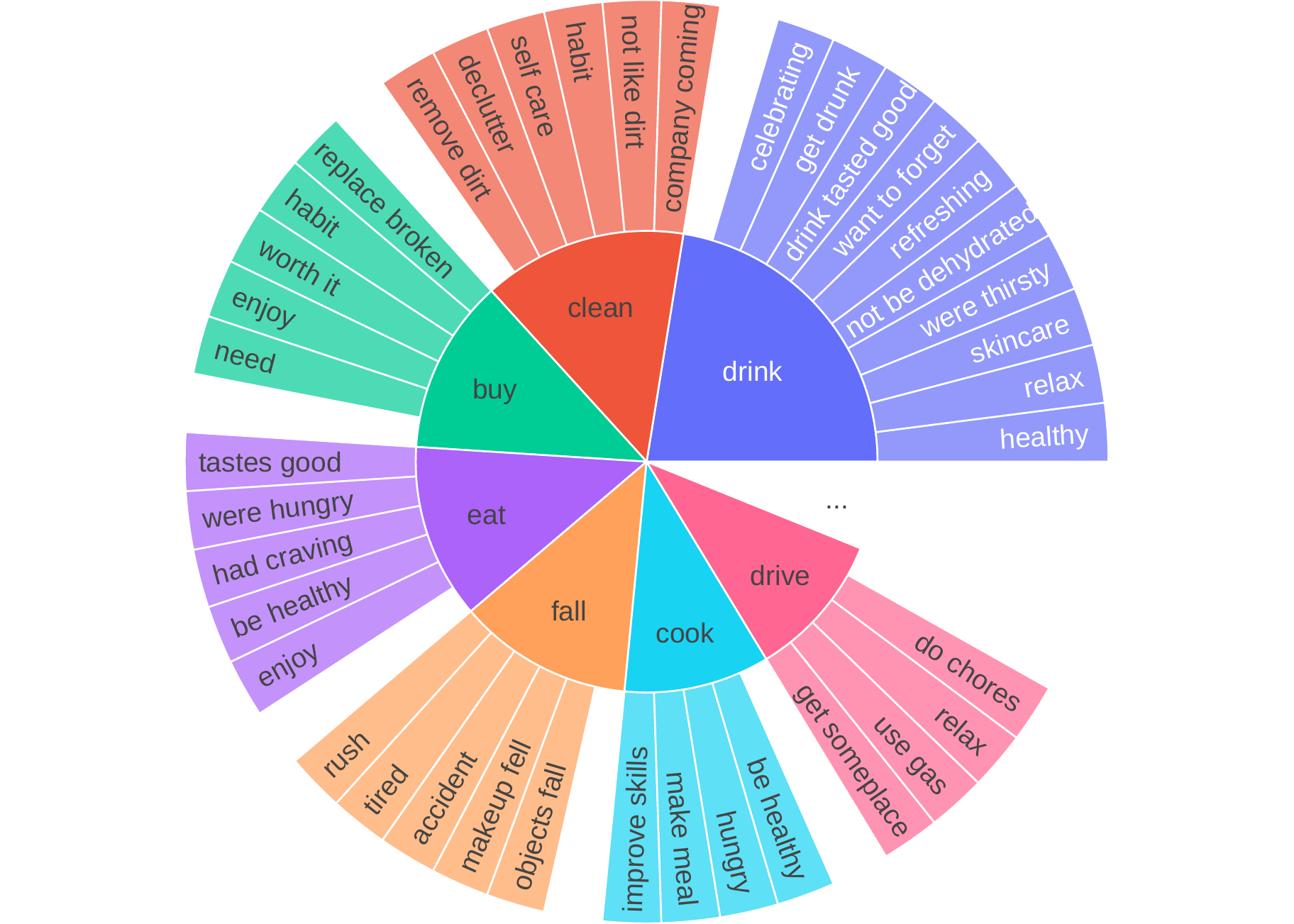}
    \caption{Distribution of the first seven actions, in alphabetical order, and their reasons, in our dataset. The rest of the actions and their reasons are shown in the appendix, in \Cref{fig:all_actions}.}
    \label{fig:some_actions}
}
\end{figure}

\begin{figure}[h]
    \centering
    \includegraphics[width=\textwidth]{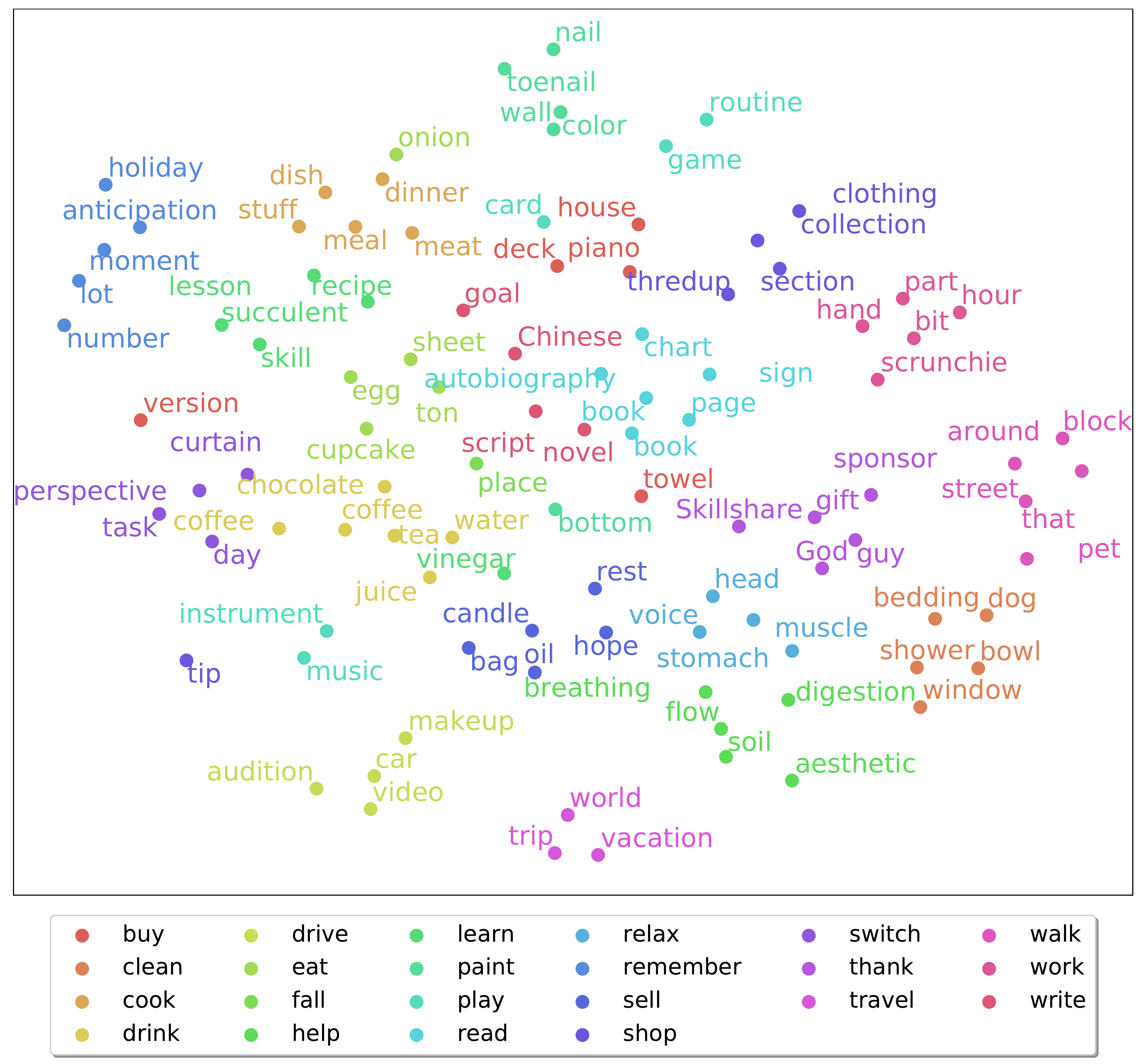}
    \caption{The t-SNE representation of the five most frequent direct objects for each action/verb in our dataset. Each color represents a different action.}
    \label{fig:cluster}
\end{figure}

\section{Data Analysis}\label{data_analysis}

We perform an analysis of the actions, reasons and video-clips in the {\sc WhyAct} dataset.
The distribution of actions and their reasons are shown in \Cref{fig:some_actions}. 
The supplemental material includes additional analyses: the distribution of actions and their number of reasons (\Cref{fig:distrib_reasons}) and videos (\Cref{fig:distrib_videos}) and the distribution of actions and their worker agreement scores (\Cref{fig:distrib_agreement}).

We also explore the content of the videos by analysing their transcripts. In particular, we look at the actions and their direct objects. For example, the action clean is depicted in various ways in the videos: ``clean shower", ``clean body", ``clean makeup", ``clean dishes".  The action diversity assures that the task is challenging and complex, trying to cover the full spectrum of everyday activities.
In \Cref{fig:cluster} we show what kind of actions are depicted in the videos: we extract all the verbs and their most five  most frequent direct objects using spaCy \cite{spacy} and then we cluster them by verb and plot them using t-distributed Stochastic Neighbor Embedding (t-SNE) \cite{Maaten08visualizingdata}.

Finally, we analyse what kind of information is required for detecting the action reasons: what is verbally described, visually shown in the video or the combination of visual and verbal cues. For this, we analyse the worker's justifications for selecting the action reasons: if the reasons were verbally mentioned in the video, visually shown or both. For each video, we take the justification selected by the majority of workers. We find that the reasons for the actions can be inferred only by relying on the narration for less than half of the videos (496 / 1,077). For the remaining videos, the annotators answered that they relied on either the visual information (in 55 videos) or on both visual and audio information (in 423 videos). The remaining 103 videos do not have a clear agreement among annotators on the modality used to indicate the action reasons.  We believe that this imbalanced split might be a reason for why the multimodal model does not perform as well as the text model. For future work, we want to collect more visual data that contains action reasons.

\paragraph{Impact of reason specificity on model performance.}
The reasons in {\sc WhyAct} vary from specific (e.g., for the verb ``fall', possible reasons are: ``tripped", ``ladder broke", ``rush", ``makeup fell") to general (e.g., for the verb ``play", possible reasons are: ``relax", ``entertain yourself", ``play an instrument"). We believe that a model can benefit from learning both general and specific reasons. From general reasons such as ``relax", a model can learn to extrapolate,  generalize, and adapt to other actions for which those reasons might apply (e.g., ``relax" can also be a reason for actions like ``drink" or ``read") and use these general reasons to learn commonalities between these actions. On the other hand, from a specific reason like ``ladder broke", the model can learn very concise even if limited information, which applies to very specific actions. 

\paragraph{Data Annotation Challenges.}
During the data annotation process, the workers had the choice to write comments about the task. From these comments we found that some difficulties with data annotation had to do with actions expressed through verbs that have multiple meanings and are sometimes used as figures of speech. For instance, the verb ``jump" was often labeled by workers as ``jumping means starting'' or ``jumping is a figure of speech here.'' Because the majority of videos  containing the verb ``jump" are labeled like this, we decided to remove this verb from our initial list of 25 actions.
Another verb that is used (only a few times) with multiple meanings is ``fall'' and some of the comments received from the workers are: ``she mentions the season fall, not the action of falling,'' ``falling is falling into place,'' ``falling off the wagon, figure of speech.'' These examples confirm how rich and complex the collected data is and how current state-of-the-art parsers are not sufficient to correctly process it.


\section{Conclusion}

In this paper, we addressed the task of detecting human action reasons in online videos. We explored the genre of lifestyle vlogs, and constructed {\sc WhyAct} -- a new dataset of 1,077 video-clips, actions and their reasons. We described and evaluated several textual and visual baselines and introduced a multimodal model that leverages both visual and textual information.

We built {\sc WhyAct} and action reason detection models to address two problems important for the advance of action recognition systems: adaptability to changing visual and textual context, and processing the richness of unscripted natural language.  
In future work, we plan to experiment with our action reason detection models in action recognition systems to improve their performance.

The dataset and the code introduced in this paper are publicly available at \url{https://github.com/MichiganNLP/vlog_action_reason}.

\section*{Ethics and Broad Impact Statement}

Our dataset contains public YouTube vlogs, in which vloggers choose to share episodes of their daily life routine. They share not only how they perform certain actions, but also their opinions and feelings about different subjects. We use the videos to detect actions and their reasons, without relying on any information about the identity of the person such as gender, age or location. 

The data can be used to better understand people's lives, by looking at their daily routine and why they choose to perform certain actions. The data contains videos of men and women and sometimes children. The routine videos present mostly ideal routines and are not comprehensive of all people's daily lives. Most of the people represented in the videos are middle class Americans.

In our data release, we only provide the YouTube urls of the videos, so the creator of the videos can always have the option to remove them. YouTube videos are a frequent source of data in research papers \cite{Miech2019HowTo100MLA,Fouhey2018FromLV,AbuElHaija2016YouTube8MAL}, and we followed the typical process used by all this previous work of compiling the data through the official YouTube API and only sharing the urls of the videos. We have the rights to use our dataset in the way we are using it, and we bear responsibility in case of a violation of rights or terms of service.

\section*{Acknowledgements}
We thank Pingxuan Huang for his help in improving the annotation user interface. This research was partially supported by a grant from the Automotive Research Center (ARC) at the University of Michigan.

\bibliography{main.bib}

\begin{thebibliography}{57}
\expandafter\ifx\csname natexlab\endcsname\relax\def\natexlab#1{#1}\fi

\bibitem[{Abu-El-Haija et~al.(2016)Abu-El-Haija, Kothari, Lee, Natsev,
  Toderici, Varadarajan, and Vijayanarasimhan}]{AbuElHaija2016YouTube8MAL}
Sami Abu-El-Haija, Nisarg Kothari, Joonseok Lee, A.~Natsev, G.~Toderici,
  Balakrishnan Varadarajan, and Sudheendra Vijayanarasimhan. 2016.
\newblock Youtube-8m: A large-scale video classification benchmark.
\newblock \emph{ArXiv}, abs/1609.08675.

\bibitem[{Angeli et~al.(2015)Angeli, Johnson~Premkumar, and
  Manning}]{Angeli2015LeveragingLS}
Gabor Angeli, Melvin~Jose Johnson~Premkumar, and Christopher~D. Manning. 2015.
\newblock \href {https://doi.org/10.3115/v1/P15-1034} {Leveraging linguistic
  structure for open domain information extraction}.
\newblock In \emph{Proceedings of the 53rd Annual Meeting of the Association
  for Computational Linguistics and the 7th International Joint Conference on
  Natural Language Processing (Volume 1: Long Papers)}, pages 344--354,
  Beijing, China. Association for Computational Linguistics.

\bibitem[{Carreira and Zisserman(2017)}]{Carreira2017QuoVA}
Joao Carreira and Andrew Zisserman. 2017.
\newblock Quo vadis, action recognition? a new model and the kinetics dataset.
\newblock In \emph{Proceedings of the IEEE Conference on Computer Vision and
  Pattern Recognition (CVPR)}.

\bibitem[{Castro et~al.(2021)Castro, Wang, Huang, Stewart, Liu, Stroud, and
  Mihalcea}]{fitb}
Santiago Castro, Ruoyao Wang, Pingxuan Huang, Ian Stewart, Nan Liu, Jonathan
  Stroud, and Rada Mihalcea. 2021.
\newblock \href {http://arxiv.org/abs/2104.04182} {Fill-in-the-blank as a
  challenging video understanding evaluation framework}.

\bibitem[{Chao et~al.(2018)Chao, Vijayanarasimhan, Seybold, Ross, Deng, and
  Sukthankar}]{Chao2018RethinkingTF}
Yu-Wei Chao, Sudheendra Vijayanarasimhan, Bryan Seybold, David~A. Ross, Jia
  Deng, and R.~Sukthankar. 2018.
\newblock Rethinking the faster r-cnn architecture for temporal action
  localization.
\newblock \emph{2018 IEEE/CVF Conference on Computer Vision and Pattern
  Recognition}, pages 1130--1139.

\bibitem[{Chen et~al.(2013)Chen, Shrivastava, and Gupta}]{Chen2013Neil}
Xinlei Chen, Abhinav Shrivastava, and Abhinav Gupta. 2013.
\newblock {NEIL}: Extracting visual knowledge from web data.
\newblock \emph{2013 IEEE International Conference on Computer Vision}.

\bibitem[{Damen et~al.(2020)Damen, Doughty, Farinella, , Furnari, Ma, Kazakos,
  Moltisanti, Munro, Perrett, Price, and Wray}]{Damen2020RESCALING}
Dima Damen, Hazel Doughty, Giovanni~Maria Farinella, , Antonino Furnari, Jian
  Ma, Evangelos Kazakos, Davide Moltisanti, Jonathan Munro, Toby Perrett, Will
  Price, and Michael Wray. 2020.
\newblock Rescaling egocentric vision.
\newblock \emph{CoRR}, abs/2006.13256.

\bibitem[{Damen et~al.(2018)Damen, Doughty, Farinella, Fidler, Furnari,
  Kazakos, Moltisanti, Munro, Perrett, Price, and Wray}]{Damen2018EPICKITCHENS}
Dima Damen, Hazel Doughty, Giovanni~Maria Farinella, Sanja Fidler, Antonino
  Furnari, Evangelos Kazakos, Davide Moltisanti, Jonathan Munro, Toby Perrett,
  Will Price, and Michael Wray. 2018.
\newblock Scaling egocentric vision: The epic-kitchens dataset.
\newblock In \emph{European Conference on Computer Vision (ECCV)}.

\bibitem[{Fang et~al.(2020)Fang, Gokhale, Banerjee, Baral, and
  Yang}]{Fang2020Video2CommonsenseGC}
Zhiyuan Fang, Tejas Gokhale, Pratyay Banerjee, Chitta Baral, and Yezhou Yang.
  2020.
\newblock \href {https://doi.org/10.18653/v1/2020.emnlp-main.61}
  {{V}ideo2{C}ommonsense: Generating commonsense descriptions to enrich video
  captioning}.
\newblock In \emph{Proceedings of the 2020 Conference on Empirical Methods in
  Natural Language Processing (EMNLP)}, pages 840--860, Online. Association for
  Computational Linguistics.

\bibitem[{Feichtenhofer et~al.(2019)Feichtenhofer, Fan, Malik, and
  He}]{Feichtenhofer2019SlowFastNF}
Christoph Feichtenhofer, Haoqi Fan, Jitendra Malik, and Kaiming He. 2019.
\newblock Slowfast networks for video recognition.
\newblock \emph{2019 IEEE/CVF International Conference on Computer Vision
  (ICCV)}, pages 6201--6210.

\bibitem[{Fleiss(1971)}]{Fleiss1971MeasuringNS}
J.~Fleiss. 1971.
\newblock Measuring nominal scale agreement among many raters.
\newblock \emph{Psychological Bulletin}, 76:378--382.

\bibitem[{Fouhey et~al.(2018)Fouhey, Kuo, Efros, and Malik}]{Fouhey2018FromLV}
David~F. Fouhey, Weicheng Kuo, Alexei~A. Efros, and Jitendra Malik. 2018.
\newblock From lifestyle vlogs to everyday interactions.
\newblock \emph{2018 IEEE/CVF Conference on Computer Vision and Pattern
  Recognition}, pages 4991--5000.

\bibitem[{Gilovich et~al.(2002)Gilovich, Griffin, and
  Kahneman}]{Gilovich2002HeuristicsAB}
Thomas Gilovich, Dale Griffin, and Daniel Kahneman. 2002.
\newblock \emph{Heuristics and biases: The psychology of intuitive judgment}.
\newblock Cambridge university press.

\bibitem[{Girdhar et~al.(2019)Girdhar, Carreira, Doersch, and
  Zisserman}]{Girdhar2019VideoAT}
Rohit Girdhar, Jo{\~a}o Carreira, Carl Doersch, and Andrew Zisserman. 2019.
\newblock Video action transformer network.
\newblock \emph{2019 IEEE/CVF Conference on Computer Vision and Pattern
  Recognition (CVPR)}, pages 244--253.

\bibitem[{He et~al.(2017)He, Lee, Lewis, and Zettlemoyer}]{He2017DeepSR}
Luheng He, Kenton Lee, Mike Lewis, and Luke Zettlemoyer. 2017.
\newblock \href {https://doi.org/10.18653/v1/P17-1044} {Deep semantic role
  labeling: What works and what{'}s next}.
\newblock In \emph{Proceedings of the 55th Annual Meeting of the Association
  for Computational Linguistics (Volume 1: Long Papers)}, pages 473--483,
  Vancouver, Canada. Association for Computational Linguistics.

\bibitem[{Hershey et~al.(2017)Hershey, Chaudhuri, Ellis, Gemmeke, Jansen,
  Moore, Plakal, Platt, Saurous, Seybold, Slaney, Weiss, and Wilson}]{45611}
Shawn Hershey, Sourish Chaudhuri, Daniel P.~W. Ellis, Jort~F. Gemmeke, Aren
  Jansen, Channing Moore, Manoj Plakal, Devin Platt, Rif~A. Saurous, Bryan
  Seybold, Malcolm Slaney, Ron Weiss, and Kevin Wilson. 2017.
\newblock \href {https://arxiv.org/abs/1609.09430} {Cnn architectures for
  large-scale audio classification}.
\newblock In \emph{International Conference on Acoustics, Speech and Signal
  Processing (ICASSP)}.

\bibitem[{Honnibal et~al.(2020)Honnibal, Montani, Van~Landeghem, and
  Boyd}]{spacy}
Matthew Honnibal, Ines Montani, Sofie Van~Landeghem, and Adriane Boyd. 2020.
\newblock \href {https://doi.org/10.5281/zenodo.1212303} {{spaCy:
  Industrial-strength Natural Language Processing in Python}}.

\bibitem[{Hwang et~al.(2021)Hwang, Bhagavatula, Le~Bras, Da, Sakaguchi,
  Bosselut, and Choi}]{Hwang2020COMETATOMIC2O}
Jena~D. Hwang, Chandra Bhagavatula, Ronan Le~Bras, Jeff Da, Keisuke Sakaguchi,
  Antoine Bosselut, and Yejin Choi. 2021.
\newblock \href {https://ojs.aaai.org/index.php/AAAI/article/view/16792}
  {(comet-) atomic 2020: On symbolic and neural commonsense knowledge graphs}.
\newblock \emph{Proceedings of the AAAI Conference on Artificial Intelligence},
  35(7):6384--6392.

\bibitem[{Iashin and Rahtu(2020)}]{BMT_Iashin_2020}
Vladimir Iashin and Esa Rahtu. 2020.
\newblock A better use of audio-visual cues: Dense video captioning with
  bi-modal transformer.
\newblock In \emph{British Machine Vision Conference (BMVC)}.

\bibitem[{Ignat et~al.(2019)Ignat, Burdick, Deng, and
  Mihalcea}]{Ignat2019IdentifyingVA}
Oana Ignat, Laura Burdick, Jia Deng, and Rada Mihalcea. 2019.
\newblock \href {https://doi.org/10.18653/v1/P19-1643} {Identifying visible
  actions in lifestyle vlogs}.
\newblock In \emph{Proceedings of the 57th Annual Meeting of the Association
  for Computational Linguistics}, pages 6406--6417, Florence, Italy.
  Association for Computational Linguistics.

\bibitem[{Jia et~al.(2021)Jia, Wu, Reiter, Cardie, Belongie, and
  Lim}]{Jia2020IntentonomyAD}
Menglin Jia, Zuxuan Wu, Austin Reiter, Claire Cardie, Serge Belongie, and
  Ser-Nam Lim. 2021.
\newblock Intentonomy: a dataset and study towards human intent understanding.
\newblock In \emph{2021 IEEE Conference on Computer Vision and Pattern
  Recognition (CVPR)}.

\bibitem[{Karpathy et~al.(2014)Karpathy, Toderici, Shetty, Leung, Sukthankar,
  and Fei-Fei}]{Karpathy2014LargeScaleVC}
A.~Karpathy, G.~Toderici, Sanketh Shetty, Thomas Leung, R.~Sukthankar, and
  Li~Fei-Fei. 2014.
\newblock Large-scale video classification with convolutional neural networks.
\newblock \emph{2014 IEEE Conference on Computer Vision and Pattern
  Recognition}, pages 1725--1732.

\bibitem[{Kong and Fu(2018)}]{Kong2018HumanAR}
Yu~Kong and Yun Fu. 2018.
\newblock Human action recognition and prediction: A survey.
\newblock \emph{ArXiv}, abs/1806.11230.

\bibitem[{Lin et~al.(2014)Lin, Maire, Belongie, Hays, Perona, Ramanan,
  Doll{\'a}r, and Zitnick}]{Lin2014MicrosoftCC}
Tsung-Yi Lin, Michael Maire, Serge Belongie, James Hays, Pietro Perona, Deva
  Ramanan, Piotr Doll{\'a}r, and C.~Lawrence Zitnick. 2014.
\newblock Microsoft coco: Common objects in context.
\newblock In \emph{Computer Vision -- ECCV 2014}, pages 740--755, Cham.
  Springer International Publishing.

\bibitem[{Miech et~al.(2019)Miech, Zhukov, Alayrac, Tapaswi, Laptev, and
  Sivic}]{Miech2019HowTo100MLA}
Antoine Miech, D.~Zhukov, Jean-Baptiste Alayrac, Makarand Tapaswi, I.~Laptev,
  and Josef Sivic. 2019.
\newblock Howto100m: Learning a text-video embedding by watching hundred
  million narrated video clips.
\newblock \emph{2019 IEEE/CVF International Conference on Computer Vision
  (ICCV)}, pages 2630--2640.

\bibitem[{Mitchell et~al.(2015)Mitchell, Cohen, Hruschka, Talukdar, Yang,
  Betteridge, Carlson, Mishra, Gardner, Kisiel, Krishnamurthy, Lao, Mazaitis,
  Mohamed, Nakashole, Platanios, Ritter, Samadi, Settles, Wang, Wijaya, Gupta,
  Chen, Saparov, Greaves, and Welling}]{Mitchell2015NeverEndingL}
Tom~Michael Mitchell, William~W. Cohen, Estevam~R. Hruschka, Partha~P.
  Talukdar, Bo~Yang, J.~Betteridge, Andrew Carlson, B.~D. Mishra, Matt Gardner,
  Bryan Kisiel, J.~Krishnamurthy, N.~Lao, Kathryn Mazaitis, Thahir Mohamed,
  Ndapandula Nakashole, Emmanouil~Antonios Platanios, Alan Ritter, M.~Samadi,
  Burr Settles, R.~C. Wang, D.~Wijaya, A.~Gupta, Xinlei Chen, Abulhair Saparov,
  Malcolm Greaves, and Joel Welling. 2015.
\newblock Never-ending learning.
\newblock \emph{Communications of the ACM}, 61:103 -- 115.

\bibitem[{Mostafazadeh et~al.(2020)Mostafazadeh, Kalyanpur, Moon, Buchanan,
  Berkowitz, Biran, and Chu-Carroll}]{Mostafazadeh2020GLUCOSEGA}
Nasrin Mostafazadeh, Aditya Kalyanpur, Lori Moon, David Buchanan, Lauren
  Berkowitz, Or~Biran, and Jennifer Chu-Carroll. 2020.
\newblock \href {https://doi.org/10.18653/v1/2020.emnlp-main.370} {{GLUCOSE}:
  {G}enera{L}ized and {CO}ntextualized story explanations}.
\newblock In \emph{Proceedings of the 2020 Conference on Empirical Methods in
  Natural Language Processing (EMNLP)}, pages 4569--4586, Online. Association
  for Computational Linguistics.

\bibitem[{Murtagh and Legendre(2014)}]{Murtagh2014WardsHA}
F.~Murtagh and P.~Legendre. 2014.
\newblock Ward’s hierarchical agglomerative clustering method: Which
  algorithms implement ward’s criterion?
\newblock \emph{Journal of Classification}, 31:274--295.

\bibitem[{Ouchi et~al.(2018)Ouchi, Shindo, and Matsumoto}]{Ouchi2018ASS}
Hiroki Ouchi, Hiroyuki Shindo, and Yuji Matsumoto. 2018.
\newblock \href {https://doi.org/10.18653/v1/D18-1191} {A span selection model
  for semantic role labeling}.
\newblock In \emph{Proceedings of the 2018 Conference on Empirical Methods in
  Natural Language Processing}, pages 1630--1642, Brussels, Belgium.
  Association for Computational Linguistics.

\bibitem[{Park et~al.(2020)Park, Bhagavatula, Mottaghi, Farhadi, and
  Choi}]{Park2020VisualCOMETRA}
Jae~Sung Park, Chandra Bhagavatula, Roozbeh Mottaghi, Ali Farhadi, and Yejin
  Choi. 2020.
\newblock Visualcomet: Reasoning about the dynamic context of a still image.
\newblock In \emph{Computer Vision -- ECCV 2020}, pages 508--524, Cham.
  Springer International Publishing.

\bibitem[{Pennington et~al.(2014)Pennington, Socher, and
  Manning}]{Pennington2014GloveGV}
Jeffrey Pennington, Richard Socher, and Christopher Manning. 2014.
\newblock \href {https://doi.org/10.3115/v1/D14-1162} {{G}lo{V}e: Global
  vectors for word representation}.
\newblock In \emph{Proceedings of the 2014 Conference on Empirical Methods in
  Natural Language Processing ({EMNLP})}, pages 1532--1543, Doha, Qatar.
  Association for Computational Linguistics.

\bibitem[{Pezzelle et~al.(2020)Pezzelle, Greco, Gandolfi, Gualdoni, and
  Bernardi}]{Pezzelle2020BeDT}
Sandro Pezzelle, Claudio Greco, Greta Gandolfi, Eleonora Gualdoni, and
  Raffaella Bernardi. 2020.
\newblock \href {https://doi.org/10.18653/v1/2020.findings-emnlp.248} {{B}e
  {D}ifferent to {B}e {B}etter! {A} {B}enchmark to {L}everage the
  {C}omplementarity of {L}anguage and {V}ision}.
\newblock In \emph{Findings of the Association for Computational Linguistics:
  EMNLP 2020}, pages 2751--2767, Online. Association for Computational
  Linguistics.

\bibitem[{Raffel et~al.(2020)Raffel, Shazeer, Roberts, Lee, Narang, Matena,
  Zhou, Li, and Liu}]{t5}
Colin Raffel, Noam Shazeer, Adam Roberts, Katherine Lee, Sharan Narang, Michael
  Matena, Yanqi Zhou, Wei Li, and Peter~J. Liu. 2020.
\newblock \href {http://jmlr.org/papers/v21/20-074.html} {Exploring the limits
  of transfer learning with a unified text-to-text transformer}.
\newblock \emph{Journal of Machine Learning Research}, 21(140):1--67.

\bibitem[{Reimers and Gurevych(2019)}]{reimers-2019-sentence-bert}
Nils Reimers and Iryna Gurevych. 2019.
\newblock \href {https://doi.org/10.18653/v1/D19-1410} {Sentence-{BERT}:
  Sentence embeddings using {S}iamese {BERT}-networks}.
\newblock In \emph{Proceedings of the 2019 Conference on Empirical Methods in
  Natural Language Processing and the 9th International Joint Conference on
  Natural Language Processing (EMNLP-IJCNLP)}, pages 3982--3992, Hong Kong,
  China. Association for Computational Linguistics.

\bibitem[{Rohrbach et~al.(2012)Rohrbach, Amin, Andriluka, and
  Schiele}]{Rohrbach2012ADF}
Marcus Rohrbach, S.~Amin, M.~Andriluka, and B.~Schiele. 2012.
\newblock A database for fine grained activity detection of cooking activities.
\newblock \emph{2012 IEEE Conference on Computer Vision and Pattern
  Recognition}, pages 1194--1201.

\bibitem[{Sap et~al.(2019)Sap, Bras, Allaway, Bhagavatula, Lourie, Rashkin,
  Roof, Smith, and Choi}]{Sap2019ATOMICAA}
Maarten Sap, Ronan~Le Bras, Emily Allaway, Chandra Bhagavatula, Nicholas
  Lourie, Hannah Rashkin, Brendan Roof, Noah~A. Smith, and Yejin Choi. 2019.
\newblock \href {https://doi.org/10.1609/aaai.v33i01.33013027} {Atomic: An
  atlas of machine commonsense for if-then reasoning.}
\newblock In \emph{AAAI}, pages 3027--3035.

\bibitem[{Shou et~al.(2017)Shou, Chan, Zareian, Miyazawa, and
  Chang}]{Shou2017CDCCN}
Zheng Shou, J.~Chan, Alireza Zareian, K.~Miyazawa, and S.~Chang. 2017.
\newblock Cdc: Convolutional-de-convolutional networks for precise temporal
  action localization in untrimmed videos.
\newblock \emph{2017 IEEE Conference on Computer Vision and Pattern Recognition
  (CVPR)}, pages 1417--1426.

\bibitem[{Sigurdsson et~al.(2017)Sigurdsson, Russakovsky, and
  Gupta}]{Sigurdsson2017WhatAA}
Gunnar~A. Sigurdsson, Olga Russakovsky, and A.~Gupta. 2017.
\newblock What actions are needed for understanding human actions in videos?
\newblock \emph{2017 IEEE International Conference on Computer Vision (ICCV)},
  pages 2156--2165.

\bibitem[{Sigurdsson et~al.(2016)Sigurdsson, Varol, Wang, Farhadi, Laptev, and
  Gupta}]{Sigurdsson2016HollywoodIH}
Gunnar~A. Sigurdsson, G{\"u}l Varol, Xiaolong Wang, Ali Farhadi, Ivan Laptev,
  and Abhinav Gupta. 2016.
\newblock Hollywood in homes: Crowdsourcing data collection for activity
  understanding.
\newblock In \emph{Computer Vision -- ECCV 2016}, pages 510--526, Cham.
  Springer International Publishing.

\bibitem[{Song et~al.(2021)Song, Ma, Sun, Yang, and Liao}]{Song2020KVLBERTKE}
Dandan Song, Siyi Ma, Zhanchen Sun, Sicheng Yang, and Lejian Liao. 2021.
\newblock \href {https://doi.org/https://doi.org/10.1016/j.knosys.2021.107408}
  {Kvl-bert: Knowledge enhanced visual-and-linguistic bert for visual
  commonsense reasoning}.
\newblock \emph{Knowledge-Based Systems}, 230:107408.

\bibitem[{Soomro et~al.(2012)Soomro, Zamir, and Shah}]{Soomro2012UCF101AD}
K.~Soomro, A.~Zamir, and M.~Shah. 2012.
\newblock Ucf101: A dataset of 101 human actions classes from videos in the
  wild.
\newblock \emph{ArXiv}, abs/1212.0402.

\bibitem[{Speer et~al.(2017)Speer, Chin, and Havasi}]{Speer2017ConceptNet5A}
Robyn Speer, Joshua Chin, and Catherine Havasi. 2017.
\newblock \href {http://aaai.org/ocs/index.php/AAAI/AAAI17/paper/view/14972}
  {Conceptnet 5.5: An open multilingual graph of general knowledge}.
\newblock In \emph{AAAI}, pages 4444--4451.

\bibitem[{Synakowski et~al.(2020)Synakowski, Feng, and
  Mart{\'i}nez}]{Synakowski2020AddingKT}
Stuart Synakowski, Q.~Feng, and A.~Mart{\'i}nez. 2020.
\newblock Adding knowledge to unsupervised algorithms for the recognition of
  intent.
\newblock \emph{International Journal of Computer Vision}, pages 1--18.

\bibitem[{Tang et~al.(2019)Tang, Ding, Rao, Zheng, Zhang, Zhao, Lu, and
  Zhou}]{Tang2019COINAL}
Yansong Tang, Dajun Ding, Yongming Rao, Yu~Zheng, Danyang Zhang, L.~Zhao, Jiwen
  Lu, and J.~Zhou. 2019.
\newblock Coin: A large-scale dataset for comprehensive instructional video
  analysis.
\newblock \emph{2019 IEEE/CVF Conference on Computer Vision and Pattern
  Recognition (CVPR)}, pages 1207--1216.

\bibitem[{Tosi(1991)}]{Tosi1991ATO}
Henry~L. Tosi. 1991.
\newblock A theory of goal setting and task performance.
\newblock \emph{Academy of Management Review}, 16:480--483.

\bibitem[{Tran et~al.(2018)Tran, Wang, Torresani, Ray, LeCun, and
  Paluri}]{Tran2018ACL}
Du~Tran, Heng Wang, L.~Torresani, Jamie Ray, Y.~LeCun, and Manohar Paluri.
  2018.
\newblock A closer look at spatiotemporal convolutions for action recognition.
\newblock \emph{2018 IEEE/CVF Conference on Computer Vision and Pattern
  Recognition}, pages 6450--6459.

\bibitem[{Van~der Maaten and Hinton(2008)}]{Maaten08visualizingdata}
Laurens Van~der Maaten and Geoffrey Hinton. 2008.
\newblock Visualizing data using t-sne.
\newblock \emph{Journal of machine learning research}, 9(11).

\bibitem[{Vaswani et~al.(2017)Vaswani, Shazeer, Parmar, Uszkoreit, Jones,
  Gomez, Kaiser, and Polosukhin}]{transformer}
Ashish Vaswani, Noam Shazeer, Niki Parmar, Jakob Uszkoreit, Llion Jones,
  Aidan~N Gomez, \L~ukasz Kaiser, and Illia Polosukhin. 2017.
\newblock \href
  {https://proceedings.neurips.cc/paper/2017/file/3f5ee243547dee91fbd053c1c4a845aa-Paper.pdf}
  {Attention is all you need}.
\newblock In \emph{Advances in Neural Information Processing Systems},
  volume~30. Curran Associates, Inc.

\bibitem[{Vondrick et~al.(2016)Vondrick, Oktay, Pirsiavash, and
  Torralba}]{Vondrick2016PredictingMO}
Carl Vondrick, Deniz Oktay, H.~Pirsiavash, and A.~Torralba. 2016.
\newblock Predicting motivations of actions by leveraging text.
\newblock \emph{2016 IEEE Conference on Computer Vision and Pattern Recognition
  (CVPR)}, pages 2997--3005.

\bibitem[{Williams et~al.(2018)Williams, Nangia, and Bowman}]{N18-1101}
Adina Williams, Nikita Nangia, and Samuel Bowman. 2018.
\newblock \href {http://aclweb.org/anthology/N18-1101} {A broad-coverage
  challenge corpus for sentence understanding through inference}.
\newblock In \emph{Proceedings of the 2018 Conference of the North American
  Chapter of the Association for Computational Linguistics: Human Language
  Technologies, Volume 1 (Long Papers)}, pages 1112--1122. Association for
  Computational Linguistics.

\bibitem[{Wu et~al.(2019)Wu, Kirillov, Massa, Lo, and
  Girshick}]{wu2019detectron2}
Yuxin Wu, Alexander Kirillov, Francisco Massa, Wan-Yen Lo, and Ross Girshick.
  2019.
\newblock Detectron2.
\newblock \url{https://github.com/facebookresearch/detectron2}.

\bibitem[{Xu et~al.(2016)Xu, Mei, Yao, and Rui}]{Xu2016MSRVTTAL}
J.~Xu, Tao Mei, Ting Yao, and Y.~Rui. 2016.
\newblock Msr-vtt: A large video description dataset for bridging video and
  language.
\newblock \emph{2016 IEEE Conference on Computer Vision and Pattern Recognition
  (CVPR)}, pages 5288--5296.

\bibitem[{Yeo et~al.(2018)Yeo, Lee, Wang, Choi, Cho, Kim~Amplayo, and
  Hwang}]{Yeo2018VisualCO}
Jinyoung Yeo, Gyeongbok Lee, Gengyu Wang, Seungtaek Choi, Hyunsouk Cho, Reinald
  Kim~Amplayo, and Seung-won Hwang. 2018.
\newblock \href {https://aclanthology.org/L18-1316} {Visual choice of plausible
  alternatives: An evaluation of image-based commonsense causal reasoning}.
\newblock In \emph{Proceedings of the Eleventh International Conference on
  Language Resources and Evaluation ({LREC} 2018)}, Miyazaki, Japan. European
  Language Resources Association (ELRA).

\bibitem[{Yin et~al.(2019)Yin, Hay, and Roth}]{Yin2019BenchmarkingZT}
Wenpeng Yin, Jamaal Hay, and Dan Roth. 2019.
\newblock \href {https://doi.org/10.18653/v1/D19-1404} {Benchmarking zero-shot
  text classification: Datasets, evaluation and entailment approach}.
\newblock In \emph{Proceedings of the 2019 Conference on Empirical Methods in
  Natural Language Processing and the 9th International Joint Conference on
  Natural Language Processing (EMNLP-IJCNLP)}, pages 3914--3923, Hong Kong,
  China. Association for Computational Linguistics.

\bibitem[{Zellers et~al.(2019)Zellers, Bisk, Farhadi, and
  Choi}]{zellers2019vcr}
Rowan Zellers, Yonatan Bisk, Ali Farhadi, and Yejin Choi. 2019.
\newblock From recognition to cognition: Visual commonsense reasoning.
\newblock In \emph{Proceedings of the IEEE/CVF Conference on Computer Vision
  and Pattern Recognition (CVPR)}.

\bibitem[{Zhang et~al.(2021)Zhang, Huo, Zhao, Song, and
  Roth}]{Zhang2020LearningCC}
Hongming Zhang, Yintong Huo, Xinran Zhao, Yangqiu Song, and Dan Roth. 2021.
\newblock Learning contextual causality between daily events from
  time-consecutive images.
\newblock In \emph{Proceedings of the IEEE/CVF Conference on Computer Vision
  and Pattern Recognition (CVPR) Workshops}, pages 1752--1755.

\bibitem[{Zhou et~al.(2018)Zhou, Xu, and Corso}]{Zhou2018TowardsAL}
Luowei Zhou, Chenliang Xu, and Jason~J. Corso. 2018.
\newblock \href
  {https://www.aaai.org/ocs/index.php/AAAI/AAAI18/paper/view/17344} {Towards
  automatic learning of procedures from web instructional videos}.
\newblock In \emph{AAAI}, pages 7590--7598.

\end{thebibliography}
\bibliographystyle{acl_natbib}

\clearpage
\newpage

\appendix

\section{Appendix}\label{sec:appendix}

\noindent\begin{minipage}{\textwidth}
    \centering
    \includegraphics[height=\textheight, width=\textwidth, keepaspectratio ]{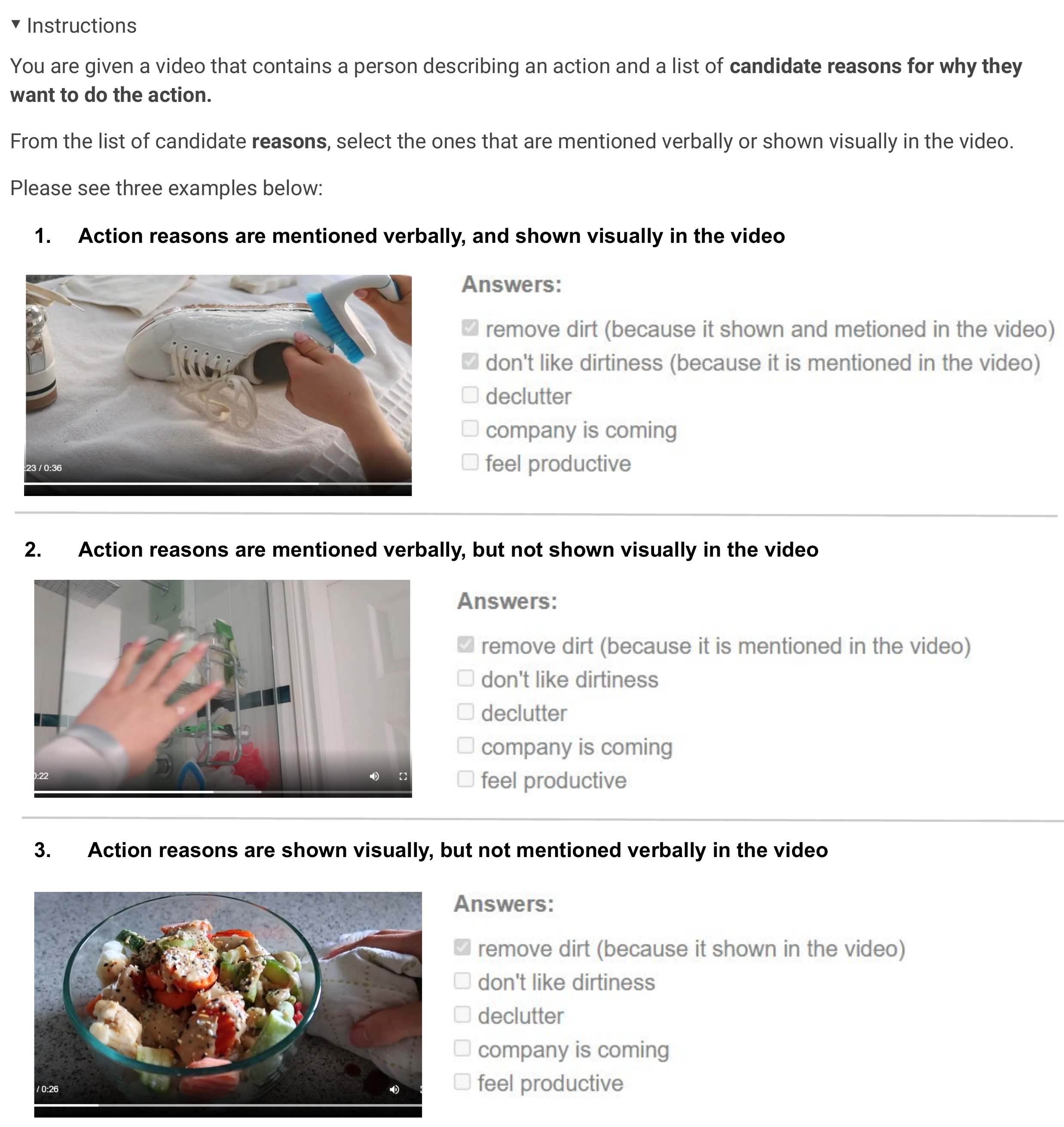}
    \captionsetup{width=0.9\linewidth}
    \captionof{figure}{Instructions and examples of completed assignments with explanations for why the answers were selected.}
    \label{fig:supp_amt}
\end{minipage}

\begin{figure*}[h]
    \centering
    \includegraphics[trim=2.3cm 0 0 0, height=15cm,keepaspectratio]{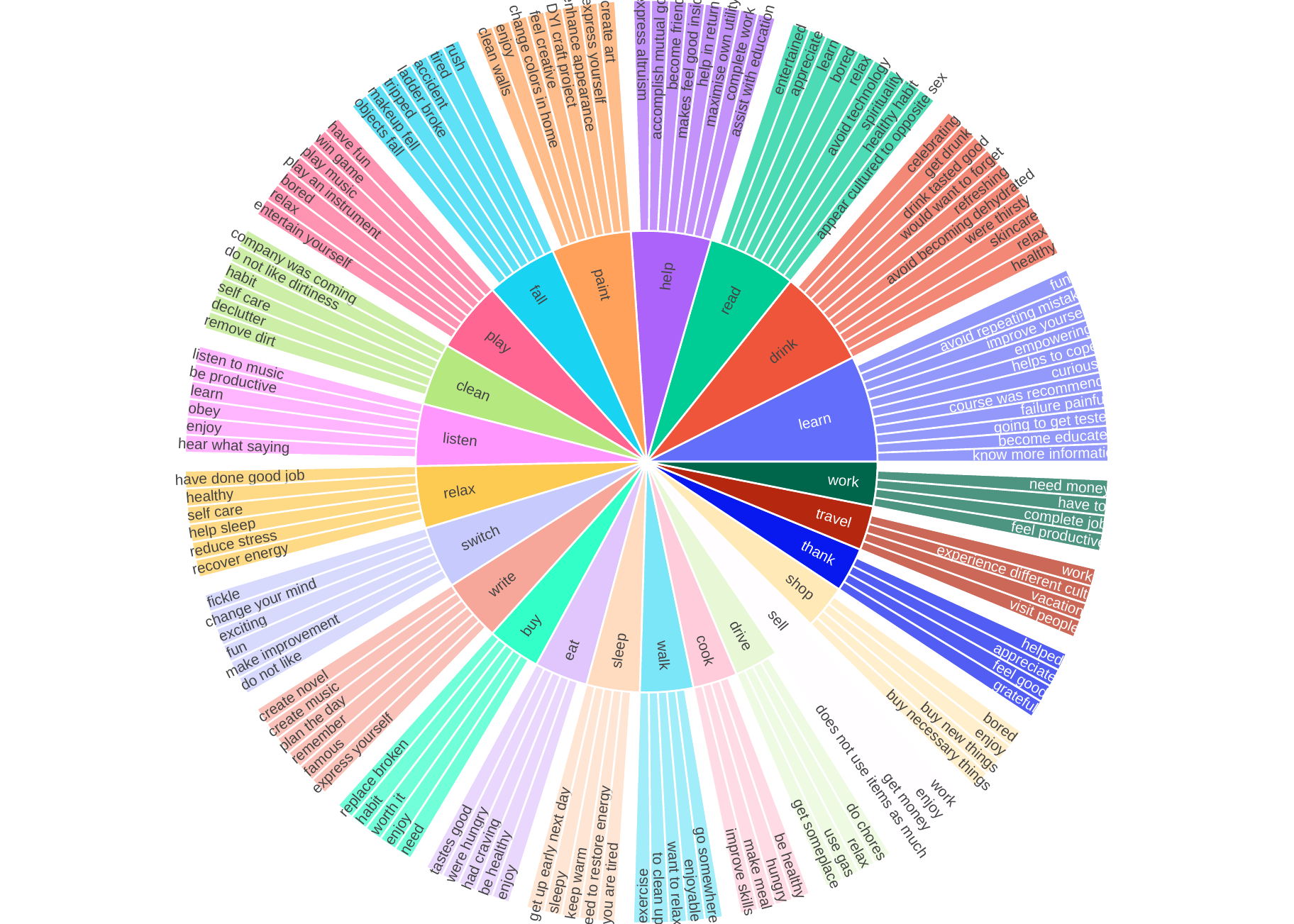}
    \caption{Distribution of all the actions and their reasons in our dataset.}
    \label{fig:all_actions}
\end{figure*}

\null
\vfill

\begin{figure*}[h]
\scalebox{0.54}{
    \centering
    \includegraphics{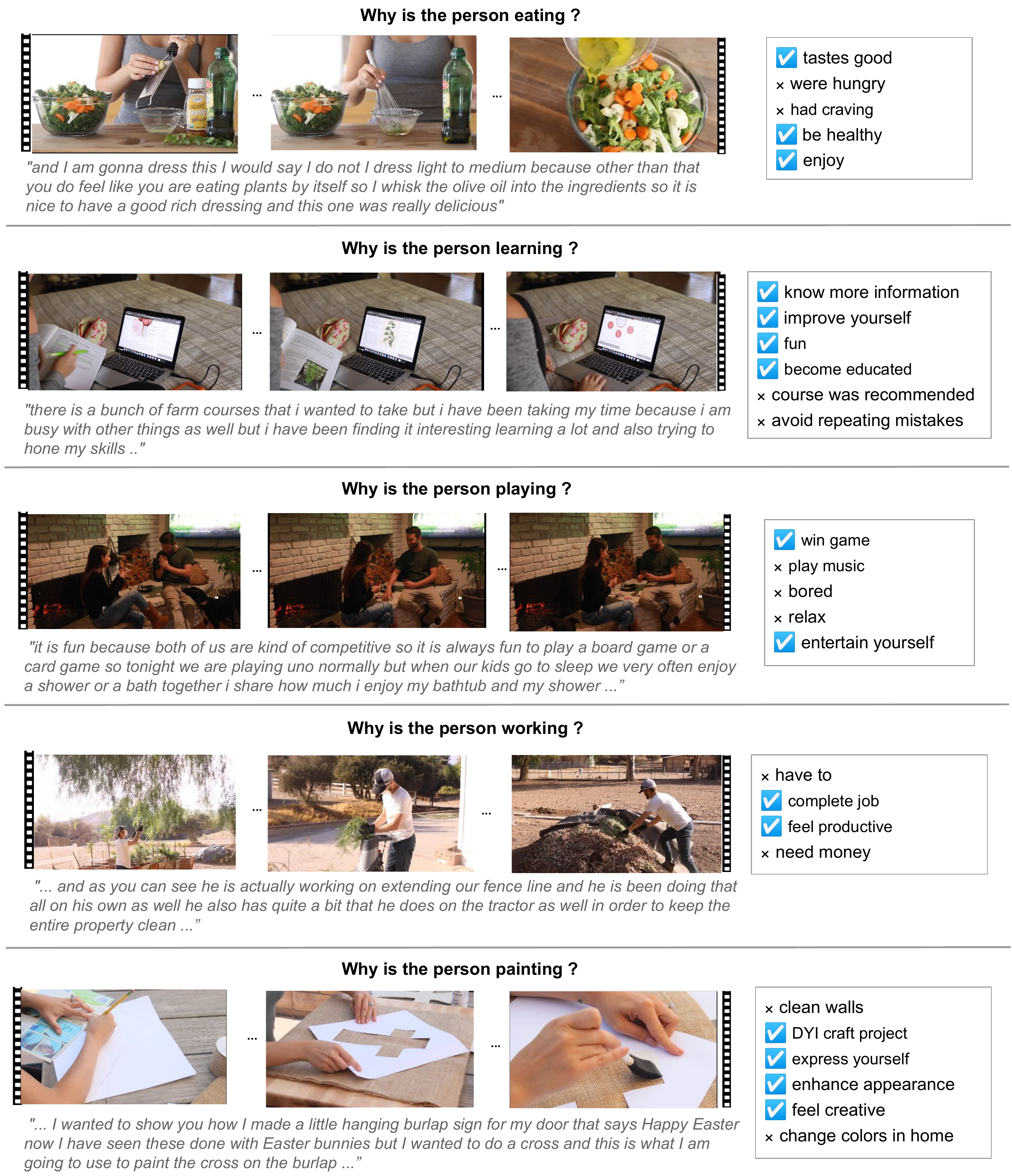}
    \caption{Other examples of actions and their annotated action reasons in our dataset.}
    \label{fig:other_examples}
    }
\end{figure*}

\begin{figure}
    \centering
    \includegraphics[width=1.2\textwidth]{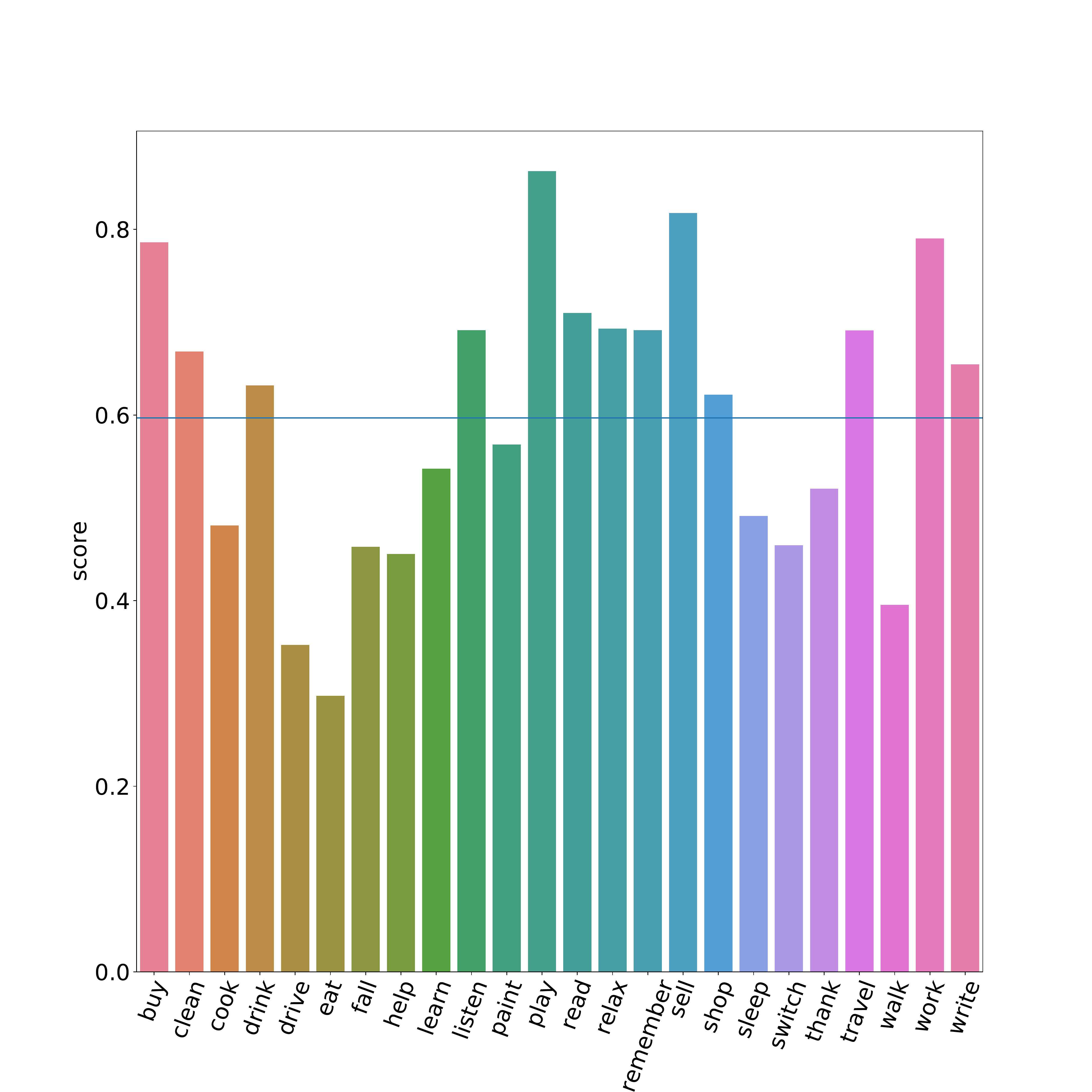}
    \caption{Distribution of all the actions and their worker agreement score: Fleiss kappa score \cite{Fleiss1971MeasuringNS}.}
    \label{fig:distrib_agreement}
\end{figure}

\begin{figure}
    \centering
    \includegraphics[width=1.1\textwidth]{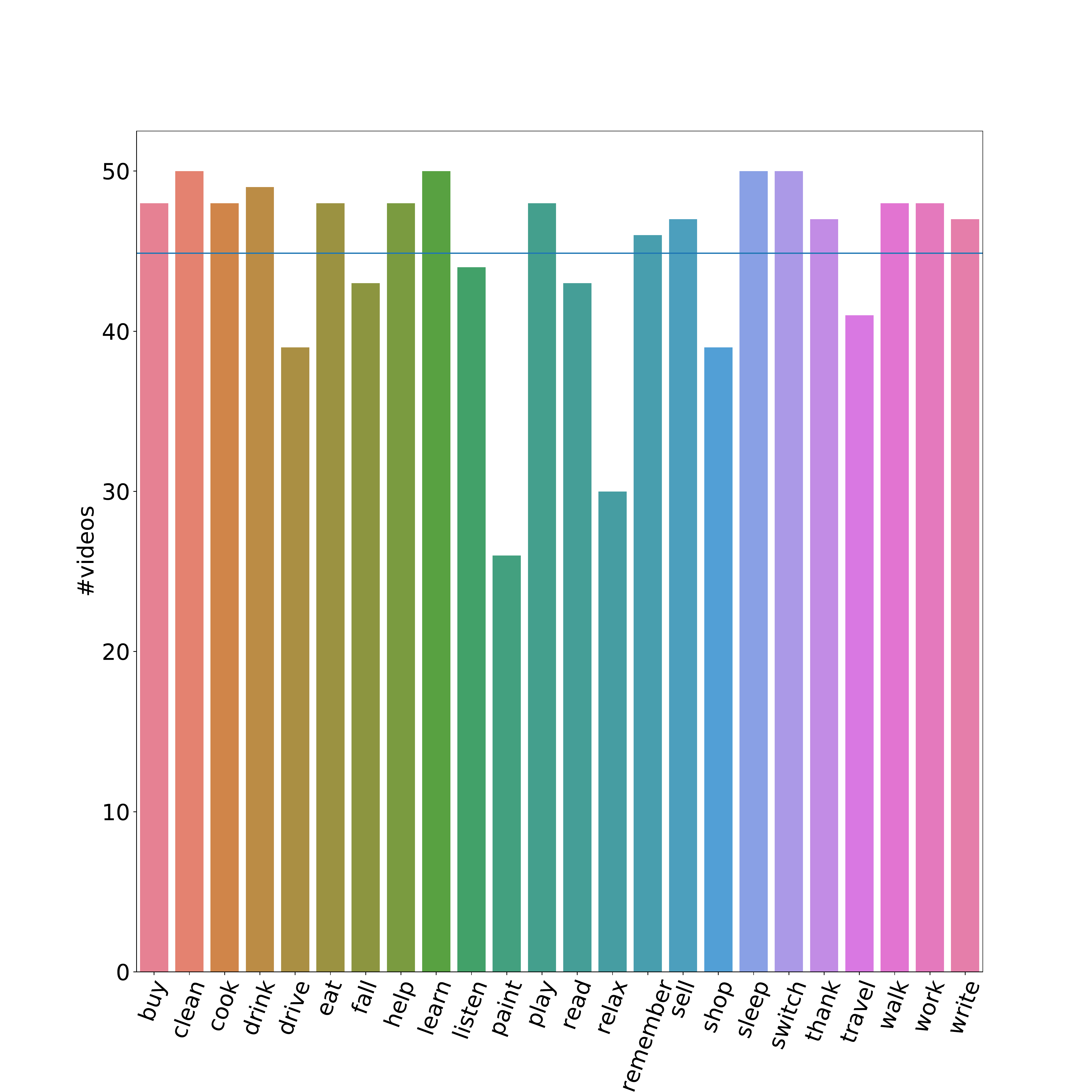}
    \caption{Distribution of all the actions and their number of videos.}
    \label{fig:distrib_videos}
\end{figure}

\begin{figure}
    \centering
    \includegraphics[width=1.1\textwidth]{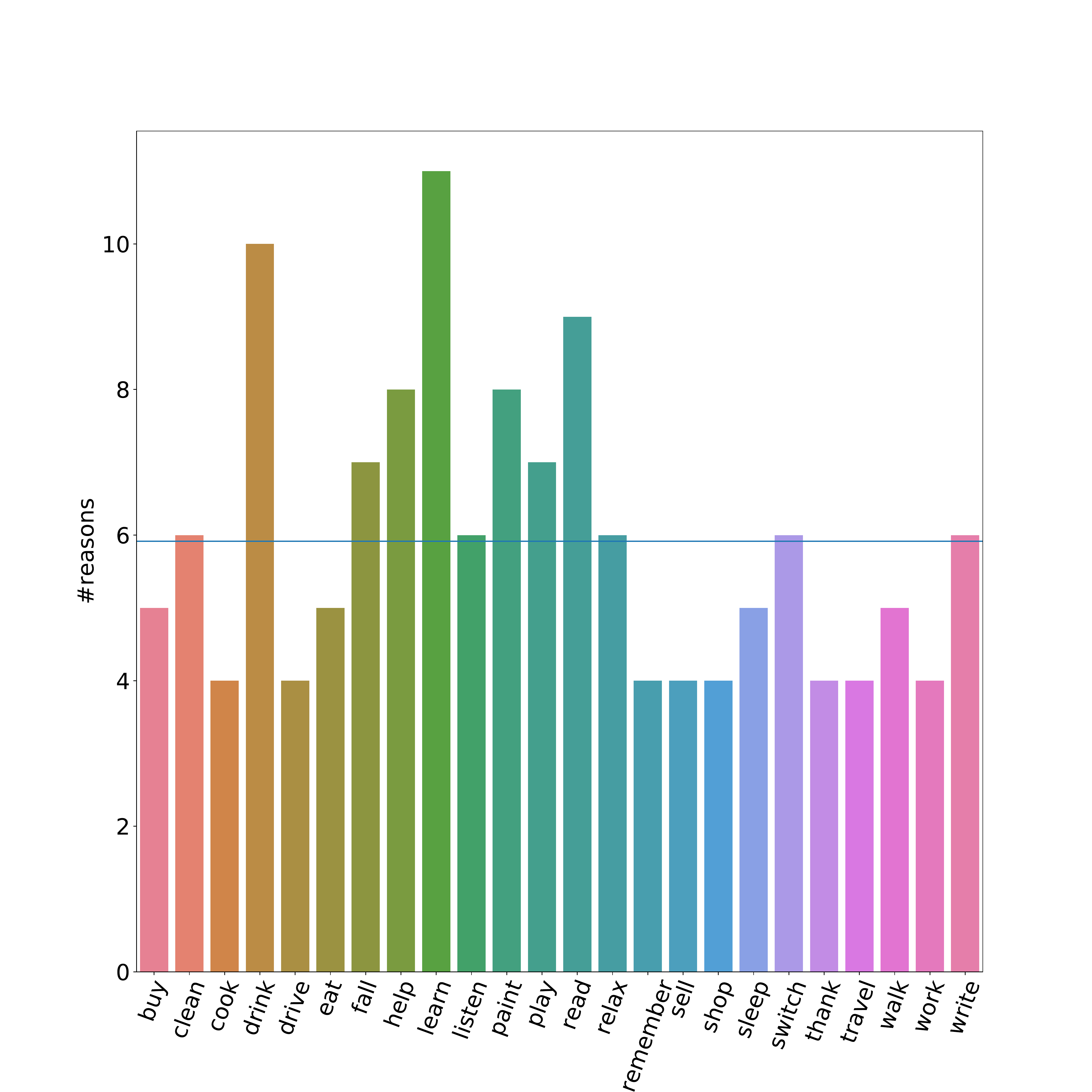}
    \caption{Distribution of all the actions and their number of reasons.}
    \label{fig:distrib_reasons}
\end{figure}

\begin{figure}
    \centering
    \includegraphics[width=1.1\textwidth]{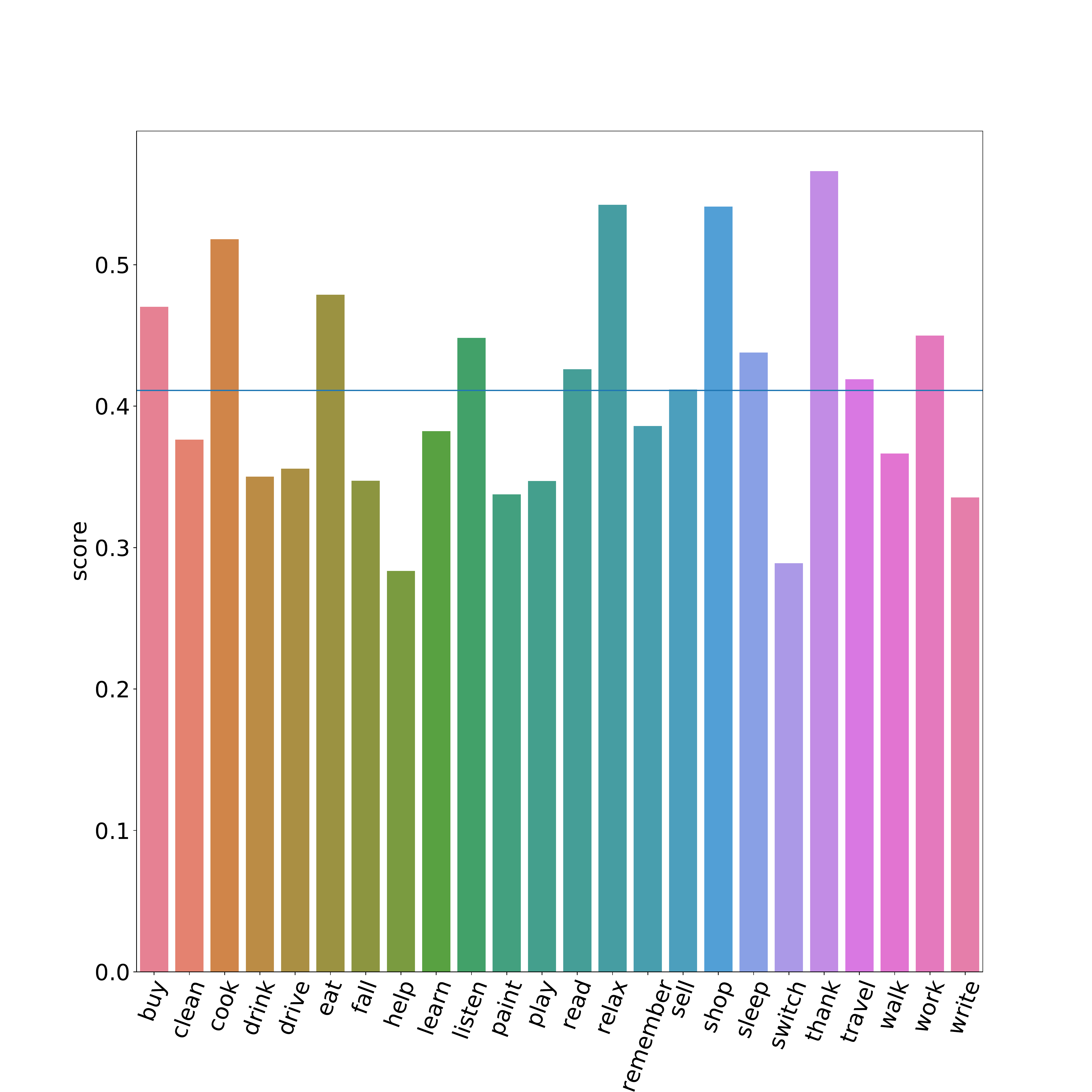}
    \caption{Distribution of all the actions and their F1 score obtained with the highest performing model (Fill-in-the-blanks with Text).}
    \label{fig:distrib_f1}
\end{figure}

\end{document}